%% file: document.tex
\documentclass[runningheads,a4paper]{llncs}

\usepackage[english]{babel}         
\usepackage[utf8]{inputenc}         
\usepackage{amsmath}                
\usepackage{amsfonts}               
\usepackage{amssymb}
\usepackage{latexsym}               
\usepackage{stmaryrd}               
\usepackage{exscale}                
\usepackage{verbatim}               
\usepackage[pdftex]{graphicx}       
\usepackage[pdftex]{hyperref}       
\usepackage{hypbmsec}               
\usepackage{listings}               
\usepackage{color}
\usepackage{array}
\usepackage{colortbl}				
\usepackage{multirow}
\usepackage{todonotes}	

\hypersetup{pdftitle={Modeling in OWL 2 without Restrictions}}
\hypersetup{pdfauthor={Michael Schneider, Sebastian Rudolph, Geoff Sutcliffe}}
\hypersetup{pdfkeywords={Semantic Web, Ontology, Modeling, OWL DL}}
\hypersetup{pdfinfo={CreationDate=D:20130428000000}}

\hypersetup{raiselinks=false} 
\hypersetup{colorlinks=false} 
\hypersetup{breaklinks=true} 

\input{macros}

\input{papervariant}

\title{Modeling in OWL 2 without Restrictions}

\author{%
Michael Schneider\inst{1}%
\and
Sebastian Rudolph\inst{2}%
\and
Geoff Sutcliffe\inst{3}%
}
\institute{%
FZI Research Center for Information Technology, Germany\\
\and
Institute AIFB, Karlsruhe Institute of Technology, Germany\\
\and
University of Miami, USA%
}

\begin{document}

\maketitle

\begin{abstract}
\input{content-abstract}

\end{abstract}

\input{content-introduction}

\input{content-preliminaries}
\input{content-usecases}
\input{content-swreasoners}

\IfPaperVariantEXTorTR
{%
\input{content-folapproach}

}%

\input{content-conclusion}

\input{content-acknowledgements}

\bibliography{biblio}
\bibliographystyle{splncs03}

\IfPaperVariantTR{%
\newpage
\appendix
\input{content-appx-testdata}

\newpage
\input{content-appx-results}
\newpage
\input{content-appx-owl2tptp}
}{}

\end{document}

%% file: macros.tex
\newcommand{\LocationSupplementaryMaterial}{%
\url{%
http://www.fzi.de/downloads/ipe/schneid/srs12-complexrelations.zip%
}
}

\newcommand{\PatternAsymmetricTransitive}{Strict Partial Orders}
\newcommand{\PatternAsymmetricChained}{Characterized Composite Relations}
\newcommand{\PatternDisjointTransitive}{Disjoint Transitive Relation Pairs}
\newcommand{\PatternDisjointChained}{Disjoint Composite Relations}
\newcommand{\PatternNumberrestrictedTransitive}{Lower-Bounded Transitive Relations}
\newcommand{\PatternNumberrestrictedChained}{Functional Composite Relations}
\newcommand{\PatternNonregularSingleChained}{Propagated Relations}
\newcommand{\PatternNonregularMultiChained}{Interlaced Relation Definitions}
\newcommand{\PatternScopedEquivalence}{Scoped Equivalence Relations}
\newcommand{\PatternReflexiveTransitive}{Quasi-Reflexive-Transitive Closures}
\newcommand{\PatternCyclicSingleRelation}{Homocyclic Relationships}
\newcommand{\PatternCyclicMultiRelation}{Heterocyclic Relationships}

\newcommand{\ACKNOWLEDGEMENTS}{\bigskip\noindent\textbf{Acknowledgements.\ }}

\newcommand{\TableFontSize}{\small}

\newcommand{\CellTextSuccess}{$\mathbf{+}$}
\newcommand{\CellColorSuccess}{\cellcolor[rgb]{0.67,1.00,0.67}}
\newcommand{\CellSuccess}{\CellTextSuccess\CellColorSuccess}

\newcommand{\CellTextWrong}{$\mathbf{-}$}
\newcommand{\CellColorWrong}{\cellcolor[rgb]{1.00,1.00,1.00}}
\newcommand{\CellWrong}{\CellTextWrong\CellColorWrong}

\newcommand{\CellTextUnknown}{$\mathbf{?}$}
\newcommand{\CellColorUnknown}{\cellcolor[rgb]{1.00,1.00,1.00}}
\newcommand{\CellUnknown}{\CellTextUnknown\CellColorUnknown}

\newcommand{\CODE}[1]{\texttt{{#1}}}
\newcommand{\TestCaseFontSize}{\scriptsize}

\newcommand{\RawResultFontSize}{\scriptsize}
\newcommand{\RawResult}[1]{{\RawResultFontSize\verbatiminput{#1}}}

\newcommand{\PaperVariantTR}{0}
\newcommand{\PaperVariantEXT}{1}
\newcommand{\PaperVariantREGULAR}{2}
\newcommand{\PaperVariantSHORT}{3}
\newcounter{PaperVariant}
\newcommand{\SetPaperVariant}[1]{\setcounter{PaperVariant}{{#1}}}
\newcommand{\IfPaperVariantREGULAR}[2]{%
	\ifthenelse{\value{PaperVariant}=\PaperVariantREGULAR}{{#1}}{{#2}}%
}
\newcommand{\IfPaperVariantEXT}[2]{%
	\ifthenelse{\value{PaperVariant}=\PaperVariantEXT}{{#1}}{{#2}}%
}
\newcommand{\IfPaperVariantTR}[2]{%
	\ifthenelse{\value{PaperVariant}=\PaperVariantTR}{{#1}}{{#2}}%
}
\newcommand{\IfPaperVariantEXTorTR}[2]{%
	\ifthenelse{{\value{PaperVariant}=\PaperVariantEXT}\or{\value{PaperVariant}=\PaperVariantTR}}{{#1}}{{#2}}%
}
\newcommand{\IfPaperVariantSHORT}[2]{%
	\ifthenelse{\value{PaperVariant}=\PaperVariantSHORT}{{#1}}{{#2}}%
}

%% file: papervariant.tex
\SetPaperVariant{\PaperVariantTR}

%% file: content-abstract.tex
The Semantic Web ontology language OWL~2 DL comes with a variety of
language features that enable sophisticated and practically useful
modeling. However, the use of these features has been severely
restricted in order to retain decidability of the language. For
example, OWL~2 DL does not allow a property to be both
transitive and asymmetric, which would be desirable, e.g., for
representing an ancestor relation. In this paper, we argue that the
so-called ``global restrictions'' of OWL~2 DL preclude many useful
forms of modeling, by providing a catalog of basic modeling patterns
that would be available in OWL~2 DL if the global restrictions were
discarded. We then report on the results of evaluating several
state-of-the-art OWL~2 DL reasoners on problems that use
combinations of features in a way that the global restrictions are
violated. The systems turn out to rely heavily on the global
restrictions and are thus largely incapable of coping with the
modeling patterns.
\IfPaperVariantEXTorTR
{%
Next we show how off-the-shelf first-order logic theorem proving technology
can be used to perform reasoning in the OWL~2 direct semantics,
the semantics that underlies OWL~2 DL,
but without requiring the global restrictions.
Applying a naive proof-of-concept implementation of this approach
to the test problems was successful in all cases.
}{}%
Based on our observations,
we make suggestions for future lines of research
on expressive description logic-style OWL reasoning.

%% file: content-introduction.tex
\section{Introduction}
\label{content-introduction}


The Semantic Web ontology language
OWL~2 DL~\cite{w3c09-owl2-document-overview,w3c09-owl2-structural-specification,w3c09-owl2-direct-semantics}
was standardized by the World Wide Web Consortium (W3C) in~2009
(and updated in 2012)
as a description logic-style formalism
of high expressivity
that still guarantees algorithmic decidability
of core reasoning tasks
such as ontology satisfiability and entailment checking.
The language
comes with a large number of language features that
enable sophisticated modeling in many application domains.
However, the use of these features
has been restricted in a variety of ways
in order to retain decidability of the language.
For instance, via a collection of \emph{``global restrictions''},
particular uses of certain features
or a combination of these features
are explicitly constrained.
A class of features
for which a large number of global restrictions have been defined
are the so-called \emph{``complex properties''},
that is,
transitive properties
and properties defined through property chain axioms.
For example,
OWL~2 DL allows for declaring transitive properties as well as asymmetric properties,
but does not allow a property to be both transitive and asymmetric,
as would be natural for a property representing an ancestor relation or, more generally,
any strict partial order.
%
%
In this paper
we analyze some of the practical ramifications
of the global restrictions on complex properties.
We argue that dropping the global restrictions
would result in a large number of
additional useful and relevant modeling options
for knowledge representation with OWL~2.
We believe that in many practical cases these advantages
outweigh the theoretical advantages of decidability.


We start in Section~\ref{content-preliminaries}
with a concise overview of OWL~2 DL 
and its global syntactic restrictions.
In Section~\ref{content-usecases} 
we provide a catalog of basic modeling patterns 
that use complex properties in natural ways, 
but which are disallowed by the global restrictions.
For all patterns, we give example use cases
supporting their usefulness and practical relevance.
For ontology authors,
the catalog offers a large set of new useful modeling options
and demonstrates the extended modeling potential available, 
in principle at least,
through the OWL~2 standard.
In the same way,
the catalog allows for a better understanding of
the actual limitations of modeling in OWL~2 DL due to the global restrictions.
To our knowledge,
no comparable catalog of patterns exists.


In Section~\ref{content-swreasoners}
we report on the results of an evaluation
of several state-of-the-art OWL~2 DL reasoners
using test problems that are based on the modeling patterns.
The evaluation results provide an understanding
of what can be expected from existing reasoners
when they are applied to input data that violates
the global restrictions of OWL~2 DL.
It has to be pointed out that such an investigation is meaningful,
as the OWL~2 standard
does \emph{not} require OWL~2 DL reasoners
to reject input beyond OWL~2 DL,
but only specifies how such tools have to behave
on legal OWL~2 DL input
(see the definition of
tool conformance for OWL~2 DL entailment checkers
in~\cite{w3c09-owl2-conformance}).
It has already been noted that OWL~2 DL reasoners can frequently be applied
to input data that is significantly beyond OWL~2 DL
without producing processing errors,
and that they sometimes produce the expected
results~\cite{schneidsut11-atp4owlfull}.
Compared to that study,
the test problems used in this paper
intuitively appear to be more digestible to OWL~2 DL reasoners.
They can be expressed in the OWL~2 structural specification~\cite{w3c09-owl2-structural-specification}
(of which OWL~2 DL is a syntactic fragment),
which has a precise meaning under the OWL~2 direct
semantics~\cite{w3c09-owl2-direct-semantics} --
the semantics underlying OWL~2 DL.
This gave rise to a hope that existing OWL~2 DL reasoners
might be able to cope with our modeling patterns.
However, the evaluation reveals
that all the OWL~2 DL reasoners
failed on all the modeling patterns.


\IfPaperVariantEXTorTR
{%
In Section~\ref{content-folapproach},
as a possible path forward,
we swiftly analyze
the practical feasibility of using 
first-order logic (FOL) reasoning technology~\cite{RV01-HAR}
to reason in OWL~2 DL without the global restrictions.
To this end
we translated our test problems into FOL theories in a
straight-forward way, following the OWL~2 direct
semantics~\cite{w3c09-owl2-direct-semantics}
but without enforcing the global restrictions of OWL~2 DL.
We then used off-the-shelf FOL theorem provers and model-finders
on the translated test data.
It turned out that this simple approach succeeded on all test cases.
Based on our observations,
we make suggestions for future lines of research
on expressive description logic-style OWL reasoning.
}{}

\IfPaperVariantTR
{%
The paper includes%
}{%
A technical report is available~\cite{techreport}
and includes%
}
\IfPaperVariantEXTorTR
{}
{%
as a possible path forward
an analysis of the
the practical feasibility of using 
first-order logic (FOL) reasoning technology~\cite{RV01-HAR}
to reason in OWL~2 DL without the global restrictions,
and gives suggestions for future lines of research
on expressive description logic-style OWL reasoning.
The technical report further includes
}%
appendices with additional technical information
on which the paper is based.
This information
\IfPaperVariantEXTorTR
{%
and the software developed for this paper
}{}%
is also available in the
\emph{supplementary material}
at
\LocationSupplementaryMaterial.

%% file: content-preliminaries.tex
\section{OWL 2 DL and Global Restrictions}
\label{content-preliminaries}

\subsection{OWL 2 DL}
\label{content-preliminaries-owl2dl}

For space reasons, we refrain from repeating the structural
specification of OWL~2 DL, and instead refer the reader to
\cite{w3c09-owl2-structural-specification} for the
complete details.
Here we focus on the aspects important
for our argument.

Recall that the basic modeling primitives in OWL are individuals,
classes and properties,
where the latter are interpreted by binary relations and strictly
subdivided into data properties and object properties. Compared to
its predecessor OWL~1, OWL~2 has been significantly extended by ways
to describe characteristics and interdependencies on the object
property level. In particular, \CODE{SubObjectPropertyOf} statements
are allowed to take \emph{property chains} as their first argument,
as, e.g., in
\begin{quote}\TestCaseFontSize
\begin{verbatim}
SubObjectPropertyOf( ObjectPropertyChain( :hasParent :hasBrother ) :hasUncle )
\end{verbatim}
\end{quote}
expressing that somebody's parent's brother is the uncle of that somebody.
In database terms one could say that the uncle relation subsumes the join
of the parent relation with the brother relation. Another novel property-centric
modeling feature is \emph{property disjointness}, as, e.g., in
\begin{quote}\TestCaseFontSize
\begin{verbatim}
DisjointObjectProperties( :hasParent :hasUncle )
\end{verbatim}
\end{quote}
expressing that somebody's parent cannot be that somebody's uncle. Furthermore,
OWL~2 allows for characterizing properties as functional, inverse-functional,
reflexive, irreflexive, symmetric, asymmetric, or transitive as, e.g., in
\begin{quote}\TestCaseFontSize
\begin{verbatim}
TransitiveObjectProperty( :hasAncestor )
\end{verbatim}
\end{quote}

Recall further that the semantics of OWL~2 DL, called \emph{direct semantics}
\cite{w3c09-owl2-direct-semantics} is established along the typical
model-theoretic semantics in description logics~\cite{baader03-dl-handbook}, and is well-defined for any
structurally specified OWL ontology even if it violates the global restrictions.

\subsection{Global Restrictions}
\label{content-preliminaries-restrictions}

In order to ensure decidability despite the high expressivity of the diverse
modeling features in OWL, the ways in which these features are allowed to
interact had to be restricted.
This led to the so-called \emph{global restrictions} 
that an OWL~2 DL ontology has to satisfy
(see Chapter~11 of~\cite{w3c09-owl2-structural-specification}). 
The name ``global restrictions'' comes
from the fact that satisfaction of these restrictions cannot be decided
by looking at the ontology axioms in isolation but it depends on their
interplay.

At the core of the restrictions is the notion of \emph{simple} versus
\emph{complex} object properties.
Roughly speaking, an object property is called complex,
if it can be inferred from the join of two or more other object properties.
For example, the above subproperty axiom renders the uncle property complex.
The same holds for the ancestor property, since transitivity of a relation
essentially means that the relation subsumes its own self-join.

The global restrictions severely constrain the ways in which complex
properties can be used: according to the \emph{restriction on simple properties},
complex properties are not allowed to occur in cardinality restrictions,
self-restrictions,
and property disjointness statements, nor is a complex property allowed to be
characterized as functional, inverse-functional, irreflexive, or asymmetric.

Another severe restriction is on the co-occurrence of subproperty axioms,
that is,
the \emph{restriction on the property hierarchy}.
The rationale behind this rather technical restriction 
is to ensure that the set of property chains used to infer a
complex property can be described as a regular language. Next to discarding
certain subproperty axioms right away, this also prohibits the coexistence of
certain such axioms.

Further restrictions apply to OWL~2 DL (referring to the use of blank nodes),
but they are not of interest in this paper.

%% file: content-usecases.tex
\section{Modeling Patterns}
\label{content-usecases}

This section presents a catalog of modeling patterns
based on usage of OWL~2 language features in a way
that violates the global restrictions of OWL~2 DL.
The catalog consists of twelve modeling patterns,
most of them representing
different combinations of
axioms defining complex properties,
such as transitivity axioms,
and language constructs that may only be used with simple properties,
such as asymmetric property axioms;
see Section~\ref{content-preliminaries} for a more detailed list 
of disallowed combinations of language constructs in OWL~2 DL.
Each modeling pattern is described
with a concrete example of a family relationship
given in OWL~2 functional syntax, and is accompanied by
an explanation of the conflicts with the OWL~2 DL specification.
Additional use cases from other application areas
provide evidence for the generality, usefulness, and relevance of the pattern.


We have to point out that the patterns were \emph{not} taken from
any existing OWL ontologies. As the patterns are explicitly
disallowed in OWL~2 DL, and as many of the involved language
features were introduced only recently as part of OWL~2, one cannot
expect to find many of these patterns in real-world ontologies
today. Rather, the goal is to demonstrate the drastic increase of
modeling power in case the global restrictions of OWL~2 DL are discarded. Our
motivations for choosing the modeling patterns were simplicity,
plausibility, potential relevance, and generality.


We are aware that for some of the patterns it is possible
to find a semantically equivalent reformulation
that is valid in OWL~2 DL.
However, the purpose of our pattern catalog is \emph{not}
to present semantic scenarios
that cannot be expressed in OWL~2 DL,
but rather to offer to ontology authors a set of new modeling options
that appear natural and simple using the features of OWL~2.
We believe that for an ontology author,
a complex or non-obvious reformulation of a pattern,
in order to keep the ontology in OWL~2 DL,
will often be unacceptable.
Still, we consider work on such translations relevant
as a means of ``repairing'' ontologies that use our modeling patterns,
so that OWL~2 DL reasoners have a better chance of coping with such input
(cf.\ the results in Section~\ref{content-swreasoners}).

\subsection{\PatternAsymmetricTransitive}
\label{content-usecases-AsymmetricTransitive}

Strict partial orders
are asymmetric transitive relations,
such as the ancestor relationship between people:
\begin{quote}\TestCaseFontSize
\begin{verbatim}
TransitiveObjectProperty( :hasAncestor )
AsymmetricObjectProperty( :hasAncestor )
\end{verbatim}
\end{quote}
OWL~2 DL does not allow transitive properties to be asymmetric.
Additional use cases include:
comparison relations such as \emph{greater-than},
part-whole relationships,
and operational research tasks such as
critical path analysis and supply chain management.

\subsection{\PatternAsymmetricChained}
\label{content-usecases-AsymmetricChained}

Property chain axioms allow composite relations to be built,
such as the uncle relation
in terms of the parent and brother relations.
Naturally, the uncle relation should be specified to be asymmetric:
\begin{quote}\TestCaseFontSize
\begin{verbatim}
SubObjectPropertyOf(
  ObjectPropertyChain( :hasParent :hasBrother )
  :hasUncle )
AsymmetricObjectProperty( :hasUncle )
\end{verbatim}
\end{quote}
OWL~2 DL does not allow composite properties to be asymmetric.
Another use case is
an asymmetric $n$\text{th}-order predecessor relation
for a fixed number~$n$,
such as a grandparent defined as a parent's parent
($n = 2$).

\subsection{\PatternDisjointTransitive}
\label{content-usecases-DisjointTransitive}

Relations are often defined as pairs
of complementary but mutually exclusive terms
that are transitive,
such as the ancestor and descendant relationships:
\begin{quote}\TestCaseFontSize
\begin{verbatim}
TransitiveObjectProperty( :hasAncestor )
TransitiveObjectProperty( :hasDescendant )
DisjointObjectProperties( :hasDescendant :hasAncestor )
\end{verbatim}
\end{quote}
OWL~2 DL does not allow disjointness of transitive properties.
Another use case is
disjoint pairs of comparison relations,
such as \emph{greater-than} and \emph{smaller-than}.
Another disallowed example of two disjoint relations
of which only one is transitive
is given by the SKOS semantic relations
\CODE{skos:broaderTransitive}
and
\CODE{skos:related}~\cite{w3c09-skos-reference} (S24, S27).

\subsection{\PatternDisjointChained}
\label{content-usecases-DisjointChained}

Relations composed using property chain axioms
are often disjoint
from one or more of the component relations.
For example,
when composing the uncle relation
in terms of the parent and brother relations,
then, realistically, all three relations are mutually disjoint:
\begin{quote}\TestCaseFontSize
\begin{verbatim}
SubObjectPropertyOf(
  ObjectPropertyChain( :hasParent :hasBrother )
  :hasUncle )
DisjointObjectProperties( :hasUncle :hasParent :hasBrother )
\end{verbatim}
\end{quote}
OWL~2 DL does not allow disjointness of composite properties.
Another use case is
an $n$\text{th}-order predecessor relation
for a fixed number~$n$,
such as a grandparent defined as a parent's parent
($n = 2$),
where the grandparent and parent relations are disjoint.

\subsection{\PatternNumberrestrictedTransitive}
\label{content-usecases-NumberrestrictedTransitive}

For some transitive relations
it may be desirable
to specify the minimum number of relationships per individual.
For example,
every person has at least two ancestors:
\begin{quote}\TestCaseFontSize
\begin{verbatim}
TransitiveObjectProperty( :hasAncestor )
SubClassOf(
  :Person
  ObjectMinCardinality( 2 :hasAncestor :Person ) )
\end{verbatim}
\end{quote}
OWL~2 DL does not allow cardinality restrictions
on transitive properties.
Other use cases are comparison relations over unbounded domains,
such as \emph{greater-than} for numbers,
where for any given number~$n$
there are always numbers~$m > n$.

\subsection{\PatternNumberrestrictedChained}
\label{content-usecases-NumberrestrictedChained}

For some relations composed by property chain axioms
it may be desirable to define them to be functional.
For example,
every person has at most one living maternal grandfather,
being the father of the person's mother:
\begin{quote}\TestCaseFontSize
\begin{verbatim}
SubObjectPropertyOf(
  ObjectPropertyChain( :hasMother :hasFather )
  :hasMaternalGrandfather )
FunctionalObjectProperty( :hasMaternalGrandfather )
\end{verbatim}
\end{quote}
OWL~2 DL does not allow composite properties to be functional.
An additional use case is a part-ownership relation,
in scenarios where items can only have a single owner
and where the owner of an item also owns all parts of the item.

\subsection{\PatternNonregularSingleChained}
\label{content-usecases-NonregularSingleChained}

Some relationships between two individuals
may ``propagate'' to two other individuals,
due to a specific constellation of relationships
that holds between all four individuals.
For example,
if Mary has mother Susan,
and Bill has father John,
where Susan and John are relatives,
then Mary and Bill are also relatives.
This can be expressed using property chain axioms:
\begin{quote}\TestCaseFontSize
\begin{verbatim}
SubObjectPropertyOf(
  ObjectPropertyChain(
    :hasMother
    :hasRelative
    ObjectInverseOf( :hasFather ) )
  :hasRelative )
\end{verbatim}
\end{quote}
This representation violates
the regularity conditions for the property hierarchy
of OWL~2 DL,
as in chains of size~3 or larger,
an inner property of the chain
(\CODE{:hasRelative} in position~2)
must not also occur as the composite property.
An additional use case would be
to characterize identical composite items,
such as computers,
to have identical corresponding components,
such as the computer's processors.

\subsection{\PatternNonregularMultiChained}
\label{content-usecases-NonregularMultiChained}

Although there is no general method in OWL~2
to fully define a composite relation,
one can sometimes encode a close characterization
by interlacing two property chain axioms.
For example,
one can define an uncle as a cousin's father,
and a cousin as an uncle's child:
\begin{quote}\TestCaseFontSize
\begin{verbatim}
SubObjectPropertyOf(
  ObjectPropertyChain( :hasCousin :hasFather )
  :hasUncle )
SubObjectPropertyOf(
  ObjectPropertyChain( :hasUncle ObjectInverseOf( :hasFather ) )
  :hasCousin )
\end{verbatim}
\end{quote}
Circular dependencies on the property hierarchy
violate the regularity conditions of OWL~2 DL.
%

\subsection{\PatternScopedEquivalence}
\label{content-usecases-ScopedEquivalence}

Equivalence relations are transitive, symmetric and reflexive,
but for reflexivity, a global scope is often not desirable,
as it would entail
that \emph{everything} has the relationship.
For example,
being a relative to someone may be seen as an equivalence relation,
provided that one accepts that everyone is a relative of himself.
However, one would probably want to limit this relation to people,
excluding, for instance, machines or ideas.
OWL~2 supports this notion of a ``locally-reflexive'' property
by self-restrictions:
\begin{quote}\TestCaseFontSize
\begin{verbatim}
SymmetricObjectProperty( :hasRelativeOrSelf )
TransitiveObjectProperty( :hasRelativeOrSelf )
EquivalentClasses( :Person ObjectHasSelf( :hasRelativeOrSelf ) )
\end{verbatim}
\end{quote}
OWL~2 DL does not allow self-restriction of transitive properties.
Other use cases include
the grouping of people according to some feature,
such as having the same profession or nationality.
SKOS specifies the mapping property \CODE{skos:exactMatch}
as symmetric and transitive~\cite{w3c09-skos-reference} (S44, S45),
and it would be plausible
and consistent with the SKOS standard
to additionally make it locally reflexive to the class of SKOS concepts.

\subsection{\PatternReflexiveTransitive}
\label{content-usecases-ReflexiveTransitive}

The reflexive-transitive closure
of a parent relation
defined over the class of people
is the smallest super relation
that is both transitive and reflexive,
where reflexivity is scoped to the class of people,
i.e., a person's ancestor or oneself.
While it is not possible to represent
the smallest such relation in OWL~2,
a coarse approximation is possible
using a self-restricted transitive super property:
\begin{quote}\TestCaseFontSize
\begin{verbatim}
SubObjectPropertyOf( :hasParent :hasAncestorOrSelf )
TransitiveObjectProperty( :hasAncestorOrSelf )
EquivalentClasses( :Person ObjectHasSelf( :hasAncestorOrSelf ) )
\end{verbatim}
\end{quote}
OWL~2 DL does not allow self-restriction of transitive properties.
Another example is \CODE{skos:broaderTransitive},
the transitive extension of the
SKOS semantic property \CODE{skos:broader}~\cite[S22,S24]{w3c09-skos-reference},
for which it would be plausible
and consistent with the SKOS standard
to additionally make it locally reflexive to the class of SKOS concepts.

\subsection{\PatternCyclicSingleRelation}
\label{content-usecases-CyclicSingleRelation}

Cyclic relationships constructed from one binary relation,
such as the ``loves'' relation,
may be of arbitrary size.
For example,
a person may love only himself
or another person mutually,
or there may be a cycle of unreturned love including several people.
Each person in such a cycle can be represented as an instance
of the class of ``loved lovers''
and, thus, instanceship in this class
indicates that a person is part of such a cyclic relationship.
Class instanceship can be expressed in terms of a
self-restricted transitive super property of the loves property:
\begin{quote}\TestCaseFontSize
\begin{verbatim}
SubObjectPropertyOf( :loves :z )
TransitiveObjectProperty( :z )
SubClassOf( ObjectHasSelf( :z ) :LovedLover )
\end{verbatim}
\end{quote}
OWL~2 DL does not allow self-restriction of transitive properties.
Another use case is
chemical ring molecules of arbitrary size,
where all bonds are of the same sort,
such as Cycloalkanes.

\subsection{\PatternCyclicMultiRelation}
\label{content-usecases-CyclicMultiRelation}

Certain cyclic relations or coincidence relations
can be composed from a set of different basic relations,
such as the concept of a legitimate child,
that is,
a person with a father and a mother who are married.
Occurrence of such relationships in a knowledge base
can be indicated by instanceship in a class of legitimate children,
modeled using a property chain axiom and a self-restriction:
\begin{quote}\TestCaseFontSize
\begin{verbatim}
SubObjectPropertyOf(
  ObjectPropertyChain(
    :hasMother
    :hasSpouse
    ObjectInverseOf( :hasFather ) )
  :z )
SubClassOf( ObjectHasSelf( :z ) :LegitimateChild )
\end{verbatim}
\end{quote}
OWL~2 DL does not allow self-restriction of composite properties.
Another use case is
circular molecules of a fixed size
that are built from different sorts of bonds,
such as Furan.

%% file: content-swreasoners.tex
\section{Evaluation of State-of-the-Art Semantic Web Reasoners}
\label{content-swreasoners}

We now report on the results of evaluating
several state-of-the-art OWL~2 DL reasoners
using test problems based on
the modeling patterns of Section~\ref{content-usecases}.
The focus was on finding out
whether or not the reasoners can cope with the modeling patterns.
As mentioned in Section~\ref{content-introduction},
compliant OWL~2 DL reasoners are \emph{not} required
to reject input beyond the specification of OWL~2 DL,
and experience shows that existing systems
often do reason upon such input.
Hence it is a legitimate question to ask
how they behave on our modeling patterns.
To give an answer,
we checked whether the reasoners are able
to recognize certain ``obvious looking'' logical conclusions
from the modeling patterns
according to the OWL~2 direct semantics.
Reasoning performance was \emph{not} considered an important aspect
of our evaluation.

\paragraph*{Test Data.}
We created a test suite
consisting of one test case
per modeling pattern.
Each test case is built from two small ontologies,
a premise and a conjecture.
The premise covers the main example
of the corresponding modeling pattern
given in Section~\ref{content-usecases},
typically extended by some additional assertions.
The conjecture is a small set of assertions
that follow logically from the premise.
Both the premise and the conjecture ontology
conform syntactically to the OWL~2 structural specification,
and the conjecture is entailed from the premise
according to the OWL~2 direct semantics.
The test cases were designed to be ``not too difficult to solve'',
so that the OWL~2 DL reasoners do not fail
due to high reasoning complexity.
The complete test suite is described in full detail
\IfPaperVariantTR
{%
in Appendix~\ref{content-appx-testdata},
}{%
in the appendix of the technical report~\cite{techreport},
}%
and in the supplementary material.

\paragraph*{Reasoners.}
We selected currently available reasoners
that represent the state of the art
in OWL~2 DL reasoning.
We used OWL API~\mbox{3.2.4} for parsing the test cases,
and all reasoners were accessed
via their respective OWL API reasoner interfaces.\footnote{%
OWL API homepage: \url{http://owlapi.sourceforge.net}%
}
\begin{itemize}

    \item
\textbf{FaCT++
\mbox{1.5.3}}
(\url{http://owl.man.ac.uk/factplusplus}),
created at the University of Manchester, England,
is a tableaux-based OWL~2 DL reasoner.

    \item
\textbf{HermiT
\mbox{1.3.6}}
(\url{http://hermit-reasoner.com}),
created at the University of Oxford, England,
is a tableaux-based OWL~2 DL reasoner.

    \item
\textbf{Pellet
\mbox{2.3.0}}
(\url{http://clarkparsia.com/pellet}),
created by Clark \& Parsia, USA,
is a tableaux-based OWL~2 DL reasoner.

\end{itemize}

\paragraph*{Testing Environment.}
All tests were conducted
on a mobile computer ``Lenovo ThinkPad T410s''
with
an Intel\textregistered~ Core\texttrademark~i5 M520 CPU (4~cores)
at 2.4~GHz speed,
4~GB RAM,
with Microsoft Windows~7 Professional (64-Bit) as the operating system.
The CPU timeout for applying a reasoner to a test case was
300~seconds.
The possible outcomes of the test runs are as follows:
\begin{itemize}
  \item `\CellTextSuccess': termination with correct result
  \item `\CellTextWrong': termination with wrong result
  \item `\CellTextUnknown': processing error or timeout
\end{itemize}

\paragraph*{Results.}
Table~\ref{tab-results-complexrelations-swreasoners}
shows the outcomes of the evaluation.
The details of the results can be found
\IfPaperVariantTR
{%
in Appendix~\ref{content-appx-results-owldlentail}
}{%
in the appendix of the technical report~\cite{techreport}
}%
and in the supplementary material.
In summary, all reasoners failed on all test cases.
\emph{FaCT++}
signalled errors on all test cases
with error messages that in most cases correctly indicated
which global restriction was violated.
\emph{HermiT}
signaled ten errors
with error messages that correctly identified
the global restriction that was violated,
while two test cases were wrongly recognized as non-entailments.
\emph{Pellet}
signaled an error in only one case,
and wrongly recognized all other test cases as non-entailments.
In the majority of cases
a warning message was found in the log file,
which explained
that Pellet had recognized a violated global restriction
and chosen to ignore one or more of the premise axioms
as a way to resolve the conflict.
Without those axioms
it was not possible to infer the conclusion ontologies.
The CPU times taken by all the reasoners
were below 20~ms for the majority of test cases,
and FaCT++ always returned after less than 10~ms.
In order to find out
whether the bad outcomes were at least partially
due to the use of the OWL API,
we compared the logical axioms and declarations
after parsing the test case data
with those in the original test cases,
and found no differences.
This indicates that the reasoners
are mainly responsible for the outcomes themselves.

\begin{table}[tb]
    \begin{center}
    \TableFontSize
    \include{includes/tab-results-complexrelations-swreasoners}
    \end{center}
    \caption{%
Entailment checking results
for the OWL~2 DL reasoners
using the twelve entailments
of the ``Complex Family Relations'' test suite.%
}
    \label{tab-results-complexrelations-swreasoners}
\end{table}

\paragraph*{Discussion.}
These results strongly indicate
that today's OWL~2 DL reasoners cannot reliably be used
with input that violates the global restrictions of OWL~2 DL.
Note that this does not mean to accuse these reasoners of malfunctioning,
they just do not go the extra mile beyond their specified input language
and hence are
not suitable for reasoning in the extended language that we are targeting.\footnote{
For someone familiar with the methods used in state-of-the-art OWL reasoners
this fact does not come as a big surprise. For instance, the restriction on the property
hierarchy is a crucial prerequisite for preprocessing the ontology ahead of
the actual core reasoning procedures.}
Apparently two different strategies are used by the reasoners
in this situation:
FaCT++ and HermiT rigidly reject the input,
while Pellet processes the input
with some of the conflicting axioms being ignored,
which may lead to missing or wrong results.
For users who want to apply some of the modeling patterns
introduced earlier in this paper,
none of these strategies is acceptable.
Therefore, a different strategy
that does not have these problems is needed.
\IfPaperVariantEXTorTR
{%
We will propose and evaluate one such strategy
in the next section.
}{%
We propose and evaluate one such strategy
in the technical report~\cite{techreport}.
}%

%% file: includes/tab-results-complexrelations-swreasoners.tex
\begin{tabular}{|l||c|c|c|c|c|c|c|c|c|c|c|c|}
\hline 
& \textbf{\textsl{01}} & \textbf{\textsl{02}} & \textbf{\textsl{03}} & \textbf{\textsl{04}} & \textbf{\textsl{05}} & \textbf{\textsl{06}} & \textbf{\textsl{07}} & \textbf{\textsl{08}} & \textbf{\textsl{09}} & \textbf{\textsl{10}} & \textbf{\textsl{11}} & \textbf{\textsl{12}} \\
\hline

\hline

\textbf{\textsl{Fact++}}
&\CellUnknown&\CellUnknown&\CellUnknown&\CellUnknown&\CellUnknown&\CellUnknown&\CellUnknown&\CellUnknown&\CellUnknown&\CellUnknown&\CellUnknown&\CellUnknown\\

\textbf{\textsl{HermiT}}
&\CellUnknown&\CellUnknown&\CellUnknown&\CellUnknown&\CellUnknown&\CellUnknown&\CellUnknown&\CellUnknown&\CellUnknown&\CellUnknown&\CellWrong&\CellWrong\\

\textbf{\textsl{Pellet}}
&\CellWrong&\CellWrong&\CellWrong&\CellWrong&\CellWrong&\CellWrong&\CellWrong&\CellWrong&\CellWrong&\CellWrong&\CellUnknown&\CellWrong\\


\hline
\end{tabular}

%% file: content-folapproach.tex
\section{First-Order-Based OWL 2 Reasoning}
\label{content-folapproach}

Based on the observation that the semantics of OWL~2 DL,
i.e., the OWL~2 direct semantics,
is based on standard first-order logic (FOL),
this section describes how
off-the-shelf FOL reasoning technology can be used for reasoning 
over the modeling patterns given in Section~\ref{content-usecases}.

\input{content-folapproach-preliminaries}
\input{content-folapproach-approach}

\input{content-folapproach-evaluation}

\input{content-folapproach-discussion}

%% file: content-folapproach-preliminaries.tex
\subsection{First-Order Logic and Automated Theorem Proving}
\label{content-folapproach-preliminaries}

Automated Theorem Proving (ATP) for classical first-order logic is a
well established field, with a solid foundation of theory that
provides the basis for the many high-performance ATP systems that
have been developed~\cite{RV01-HAR}. ATP for classical first-order
logic is semi-decidable, i.e., given axioms and a provable
conjecture it is possible to build an ATP system that, in principle,
is guaranteed to prove that the conjecture is a theorem logically
entailed by the axioms. On the flip-side, given a satisfiable set of
first-order formulae, it is not possible in general to build an ATP
system that can confirm their satisfiability, e.g., by finding a
model of the formulae. In practice, many ATP systems sacrifice
theoretical completeness of theorem proving for better practical
completeness over established work profiles, and the theoretical
incompleteness of model finding systems has not prevented them from
being practically useful in many ways. ATP has been successfully
used in a broad range of application domains, such as mathematics,
hardware and software verification, social sciences, agents and
planning, and, as is relevant to this paper, reasoning for the
Semantic Web. In concrete daily practice, many ATP systems use the
TPTP language~\cite{sutcliffe10-tptp-technical-manual} for problem
input and solution output, and the TPTP language has been used for
this work.

%% file: content-folapproach-approach.tex
\subsection{FOL-Based OWL 2 Reasoning Approach}
\label{content-folapproach-approach}

The reasoning approach consists of two steps.
Firstly,
OWL~2 ontologies are translated into TPTP formulae
following the OWL~2 direct semantics.
Secondly, FOL reasoning systems are applied to the TPTP formulae.
%
%
The translation of OWL~2 ontologies into TPTP closely follows the
definitions in the OWL~2 direct semantics
specification~\cite{w3c09-owl2-direct-semantics}, which give
semantic meaning to all OWL~2 axioms and expressions using a
FOL/set-theoretical model theory. For instance, the example OWL~2
axiom from Section~\ref{content-preliminaries-owl2dl}
\begin{quote}\TestCaseFontSize
\begin{verbatim}
SubObjectPropertyOf( ObjectPropertyChain( :hasParent :hasBrother ) :hasUncle )
\end{verbatim}
\end{quote}
has the following meaning
according to Table~6 of the OWL~2 direct semantics:
{\TestCaseFontSize
\[
\begin{array}{ll}
\forall y_0, y_1, y_2:
    ( y_0 , y_1 ) \in \texttt{:hasParent}^{\textsl{OP}}
    \wedge
    ( y_1 , y_2 ) \in \texttt{:hasBrother}^{\textsl{OP}}
    \Rightarrow
    ( y_0 , y_2 ) \in \texttt{:hasUncle}^{\textsl{OP}}
\end{array}
\]
}
and is mapped into the following TPTP formula:
\begin{quote}\TestCaseFontSize
\begin{verbatim}
! [Y0,Y1,Y2] : ( ( uri_hasParent(Y0,Y1) & uri_hasBrother(Y1,Y2) ) => uri_hasUncle(Y0,Y2) )
\end{verbatim}
\end{quote}
An implementation of the translation is available as an executable
tool in the supplementary material.


For satisfiability checking,
the ontology to be checked is translated into a TPTP axiom formula.
For entailment checking,
the premise ontology is translated into a TPTP axiom formula,
and the conclusion ontology is translated into a TPTP conjecture formula.
The TPTP formulae are then given to FOL reasoning systems,
typically a theorem prover and a model-finder,
ideally applied to the input data in parallel.
The theorem prover will try to detect
that the input is unsatisfiable or an entailment,
while the model-finder will try to detect
that the input is satisfiable or a non-entailment.


This idea has been used in the past for reasoning experiments 
in OWL~1 DL~\cite{tsarkov04-hoolet}.
Those results are now largely outdated and are not very representative,
as they use only a single FOL theorem prover (Vampire)
in a very old version,
and no FOL model-finders were used for model-finding tasks.
Moreover, the reported experiments were restricted 
to input data strictly in the scope of OWL~1 DL,
while most of our modeling scenarios are only possible in OWL~2
and they are all outside the scope of OWL~2 DL.
Recently there have also been
experiments applying FOL reasoning
to OWL~2 Full~\cite{schneidsut11-atp4owlfull}.
However, 
the approach described there does not apply to our work,
as OWL~2 Full uses a different semantics to OWL~2 DL,
and as the approach used there differs strongly from our approach
by the use of a FOL axiomatisation of the semantics of OWL~2 Full
and by translating input ontologies according to their RDF graph representation
instead of their representation and semantic meaning as OWL constructs.

%% file: content-folapproach-evaluation.tex
\subsection{Evaluation}
\label{content-folapproach-evaluation}

We now present the results of evaluating
the FOL-based reasoning approach
using several state-of-the-art FOL reasoners.
As test data,
we used the ``Complex Family Relations''
that was used for the evaluation of the OWL~2 DL reasoners
in Section~\ref{content-swreasoners}.
The test cases
were translated into TPTP formulae 
as described in Section~\ref{content-folapproach-approach},
and given to the FOL reasoners for entailment checking.
We also applied the FOL reasoners 
to the (provably satisfiable) premise ontologies of the test cases
to check whether FOL reasoners can be used
for the generally undecidable task of satisfiability checking
in the unrestricted OWL~2 direct semantics.

%
\paragraph*{The Reasoners.}
The following FOL theorem provers and model finders
were used in the evaluation,
using their current stable version at the time of writing:
\begin{itemize}

	\item
\textbf{E
\mbox{1.5}}
(\url{http://www.eprover.org}) \cite{Sch02},
created 
at the Technische Universit{\"a}t M{\"u}nchen, Germany,
is a purely equational theorem prover, using a saturation 
algorithm that implements an instance of the superposition calculus with 
negative literal selection.

	\item
\textbf{iProver
\mbox{0.9.2}}
(\url{http://www.cs.man.ac.uk/~korovink/iprover}) \cite{Kor08},
created 
at the University of Manchester, England,
is based on the Inst-Gen calculus. 
It combines ordered resolution with ground reasoning.

	\item
\textbf{SPASS
\mbox{3.5}}
(\url{http://www.spass-prover.org}) \cite{WF+09},
created 
at the Max-Planck-Institut f{\"u}r Informatik, Germany,
is a saturation based theorem prover, employing superposition, 
sorts and splitting.

	\item
\textbf{Vampire
\mbox{2.5}}
(\url{http://www.vprover.org}) \cite{RV02},
created 
at the University of Manchester, England,
is a theorem prover implementing the calculi of ordered 
binary resolution, superposition, and the Inst-gen calculus. 
Strategy scheduling is used to apply different combinations of techniques.

	\item
\textbf{Paradox
\mbox{4.0}}
(\url{http://www.cse.chalmers.se/~koen/code/}) \cite{CS03},
created 
at Chalmers University of Technology, Sweden,
is a finite model finder, based on MACE-style 
flattening and instantiation, and the use of a SAT solver to solve the 
resulting problem.

	\item
\textbf{DarwinFM
\mbox{1.4.5}}
(\url{http://goedel.cs.uiowa.edu/Darwin}) \cite{BF+06},
created 
at NICTA, Australia, and the University of Iowa, USA,
is a finite model finder in the spirit of Paradox.
For each domain size the problem is transformed into an equisatisfiable 
function-free clause set, which is decided by the Darwin prover \cite{BFT06}.

\end{itemize}

\paragraph*{Testing Environment.}
We used the TPTP reasoning service~\cite{Sut10},
available at
\url{http://tptp.org/cgi-bin/SystemOnTPTP},
which offers online access to all the FOL reasoners.
Readers may use this service
to repeat the experiments conducted here.
The underlying machine has
4~Intel\textregistered~ Xeon\textregistered~ 5140 CPUs 
at 2.33~GHz speed,
4~GB RAM (1~GB per CPU),
with Linux~2.6.31 as the operating system.
%
The possible outcomes were the same as described in 
Section~\ref{content-swreasoners}.
The CPU timeout was again set to 300~seconds.

\paragraph*{Results.}

Tables \ref{tab-results-complexrelations-folreasoners-posentail}
and~\ref{tab-results-complexrelations-folreasoners-satisfy}
show the outcomes of the 
entailment checking and satisfiability checking evaluation,
respectively.
The details of the results can be found 
\IfPaperVariantTR
{%
in the Appendices 
\ref{content-appx-results-folatpentail} 
and~\ref{content-appx-results-folatpsat}
}{%
in the appendix of the technical report~\cite{techreport}
}
and in the supplementary material.
No wrong results were produced by any reasoner.
All theorem provers 
(listed in the upper part of the tables)
succeeded on all entailment tests,
and all model finders
(listed in the lower part of the tables)
succeeded on all satisfiability tests.
Hence, any pair of a theorem prover and a model finder applied in parallel
would succeed on all the test cases.
The reasoning times were always below 50~ms
and in many cases below 10~ms,
so they are comparable 
with those of the OWL 2 DL reasoners reported
in Section~\ref{content-swreasoners}.
While these times are of limited explanatory power
(given the small sizes of the test data),
we can at least see 
that the FOL reasoners had no difficulty
with this input data.
It is also interesting to observe
that all the theorem provers succeeded on most of the satisfiability tests,
and all the model finders succeeded on most of the entailment tests.
In fact, each model finder terminated 
on the majority of entailment test cases
in less then 10~ms.
The few cases of ``unknown'' outcomes in the tables were timeouts,
i.e., no explicit error was ever reported.

\begin{table}[tb]
	\begin{center}
	\TableFontSize
	\include{includes/tab-results-complexrelations-folreasoners-posentail}
	\end{center}
	\caption{%
Entailment checking results 
for the FOL reasoning systems
using the twelve entailments
of the ``Complex Family Relations'' test suite.%
}
	\label{tab-results-complexrelations-folreasoners-posentail}
\end{table}
 
\begin{table}[tb]
	\begin{center}
	\TableFontSize
	\include{includes/tab-results-complexrelations-folreasoners-satisfy}
	\end{center}
	\caption{%
Satisfiability checking results 
for the FOL reasoning systems
using the twelve premise ontologies
of the ``Complex Family Relations'' test suite.%
}
	\label{tab-results-complexrelations-folreasoners-satisfy}
\end{table}

%% file: includes/tab-results-complexrelations-folreasoners-posentail.tex
\begin{tabular}{|l||c|c|c|c|c|c|c|c|c|c|c|c|}
\hline 
& \textbf{\textsl{01}} & \textbf{\textsl{02}} & \textbf{\textsl{03}} & \textbf{\textsl{04}} & \textbf{\textsl{05}} & \textbf{\textsl{06}} & \textbf{\textsl{07}} & \textbf{\textsl{08}} & \textbf{\textsl{09}} & \textbf{\textsl{10}} & \textbf{\textsl{11}} & \textbf{\textsl{12}} \\
\hline

\hline

\textbf{\textsl{E}}
&\CellSuccess&\CellSuccess&\CellSuccess&\CellSuccess&\CellSuccess&\CellSuccess&\CellSuccess&\CellSuccess&\CellSuccess&\CellSuccess&\CellSuccess&\CellSuccess\\

\textbf{\textsl{iProver}}
&\CellSuccess&\CellSuccess&\CellSuccess&\CellSuccess&\CellSuccess&\CellSuccess&\CellSuccess&\CellSuccess&\CellSuccess&\CellSuccess&\CellSuccess&\CellSuccess\\

\textbf{\textsl{SPASS}}
&\CellSuccess&\CellSuccess&\CellSuccess&\CellSuccess&\CellSuccess&\CellSuccess&\CellSuccess&\CellSuccess&\CellSuccess&\CellSuccess&\CellSuccess&\CellSuccess\\

\textbf{\textsl{Vampire}}
&\CellSuccess&\CellSuccess&\CellSuccess&\CellSuccess&\CellSuccess&\CellSuccess&\CellSuccess&\CellSuccess&\CellSuccess&\CellSuccess&\CellSuccess&\CellSuccess\\

\hline

\textbf{\textsl{DarwinFM}}
&\CellSuccess&\CellSuccess&\CellSuccess&\CellSuccess&\CellUnknown&\CellSuccess&\CellSuccess&\CellSuccess&\CellSuccess&\CellSuccess&\CellSuccess&\CellSuccess\\

\textbf{\textsl{Paradox}}
&\CellSuccess&\CellSuccess&\CellSuccess&\CellSuccess&\CellUnknown&\CellSuccess&\CellSuccess&\CellSuccess&\CellSuccess&\CellSuccess&\CellSuccess&\CellSuccess\\

\hline
\end{tabular}

%% file: includes/tab-results-complexrelations-folreasoners-satisfy.tex
\begin{tabular}{|l||c|c|c|c|c|c|c|c|c|c|c|c|}
\hline 
& \textbf{\textsl{01}} & \textbf{\textsl{02}} & \textbf{\textsl{03}} & \textbf{\textsl{04}} & \textbf{\textsl{05}} & \textbf{\textsl{06}} & \textbf{\textsl{07}} & \textbf{\textsl{08}} & \textbf{\textsl{09}} & \textbf{\textsl{10}} & \textbf{\textsl{11}} & \textbf{\textsl{12}} \\
\hline

\hline

\textbf{\textsl{E}}
&\CellSuccess&\CellSuccess&\CellSuccess&\CellSuccess&\CellUnknown&\CellSuccess&\CellSuccess&\CellSuccess&\CellSuccess&\CellSuccess&\CellSuccess&\CellSuccess\\

\textbf{\textsl{iProver}}
&\CellSuccess&\CellSuccess&\CellSuccess&\CellSuccess&\CellSuccess&\CellSuccess&\CellSuccess&\CellSuccess&\CellSuccess&\CellSuccess&\CellSuccess&\CellSuccess\\

\textbf{\textsl{SPASS}}
&\CellSuccess&\CellSuccess&\CellSuccess&\CellSuccess&\CellUnknown&\CellSuccess&\CellSuccess&\CellSuccess&\CellSuccess&\CellSuccess&\CellSuccess&\CellSuccess\\

\textbf{\textsl{Vampire}}
&\CellSuccess&\CellSuccess&\CellSuccess&\CellSuccess&\CellUnknown&\CellSuccess&\CellSuccess&\CellSuccess&\CellSuccess&\CellSuccess&\CellSuccess&\CellSuccess\\

\hline

\textbf{\textsl{DarwinFM}}
&\CellSuccess&\CellSuccess&\CellSuccess&\CellSuccess&\CellSuccess&\CellSuccess&\CellSuccess&\CellSuccess&\CellSuccess&\CellSuccess&\CellSuccess&\CellSuccess\\

\textbf{\textsl{Paradox}}
&\CellSuccess&\CellSuccess&\CellSuccess&\CellSuccess&\CellSuccess&\CellSuccess&\CellSuccess&\CellSuccess&\CellSuccess&\CellSuccess&\CellSuccess&\CellSuccess\\

\hline
\end{tabular}

%% file: content-folapproach-discussion.tex
\subsection{Discussion}
\label{content-folapproach-discussion}

Research in description logic-style OWL reasoning has focused
largely on creating feature-rich but decidable ontology languages,
and reasoners that closely conform to the specification of the
language and its usage. We have seen that this approach prohibits
many practically useful modeling options, and that current OWL~2 DL
reasoners are not able to go the extra mile and cope with such
modeling options. We have also seen that these scenarios can be
dealt with in a straightforward manner using FOL reasoning technology.
FOL reasoning is undecidable,
and it can probably not compete in performance and scalability
with highly specialized description-logic-style OWL~2 DL reasoning
on OWL 2 DL input.
However,
due to the compatibility of the semantics of OWL~2 DL and FOL,
it will be possible to build hybrid systems
from both kinds of reasoning systems.
An inexpensive syntax check can be used
to direct all valid OWL 2 DL input to an OWL~2 DL system,
and input beyond OWL 2 DL to a FOL system.

The hybrid approach is fully complete on valid OWL~2 DL input,
and complete with regard to entailment and unsatisfiability \emph{detection}
even outside OWL~2 DL,
due to the \emph{semi}-decidability of FOL reasoning.
Completeness is further guaranteed
for non-entailment and satisfiability detection
in cases where a finite model exists.
Regarding the remaining undecidable task
of detecting satisfiability if no finite model exists,
it will be necessary to investigate
how frequently this case occurs in practice
(it did not occur in any of our tests).
Existing FOL reasoners will sometimes be able to succeed even in these
cases,
and tools such as Infinox~\cite{CL09} can be used to detect cases where
no finite model exists,
and one can also try to develop specialized incomplete
methods for finding infinite models
to cover the practically most relevant scenarios of unrestricted OWL reasoning.

While the performance of a hybrid system applied to legal OWL~2 DL input
would presumably be comparable with that of existing OWL~2 DL reasoners
(the only difference is an additional syntax check),
we would also want its performance
to be acceptable outside the borders of OWL 2 DL.
Our results from the small test problems in our evaluation
do not allow us
to make a general statement about performance.
As mentioned in Section~\ref{content-folapproach-preliminaries},
ATP has been successfully applied to a broad range of real-world applications,
so one might be confident about their performance in OWL reasoning as well.
However,
current FOL reasoners do not provide any specific support for OWL features,
and our proof-of-concept system did not include
any OWL-specific optimizations.
It will therefore be advisable
to apply this approach to larger test cases,
and to spend considerable research effort 
in developing optimization methods for FOL-based OWL reasoning.

The hybrid reasoning strategy will allow description-logic researchers
to continue their successful work of further optimizing reasoning performance
in the case of fully compliant OWL~2~DL input,
while new research possibilities will open up
concerning the optimization of ATP-based OWL reasoning,
even including extensions beyond the current limits of OWL,
such as Boolean property expressions,
or rule-style extensions (SWRL, RIF+OWL DL combinations).

%% file: content-conclusion.tex
\section{Conclusion}
\label{content-conclusion}

In this paper
we have shown that
many useful and relevant modeling options
were available in OWL~2 DL if the global restrictions on complex properties
would be relinquished.
We have presented a catalog of twelve basic useful modeling patterns
that are in the scope of the 
unrestricted structural specification 
and the direct semantics of OWL~2
but are beyond the scope of OWL~2 DL,
including strict partial orders and different forms of circular relationships.
Although the OWL~2 DL standard does not prevent compliant reasoners
from processing such input,
all the state-of-the-art OWL~2 DL reasoners that we tested
were unable to cope with these modeling scenarios.

\IfPaperVariantEXTorTR
{%
The use of generic FOL reasoning technology
}{%
In our technical report~\cite{techreport}
we have analysed the use of generic FOL reasoning technology
for reasoning in the unrestricted OWL~2 direct semantics,
and our experiments
}%
led to fully satisfying results on our test data.
We therefore suggest building loosely coupled hybrid OWL~2 reasoners
from traditional tableaux-based systems and generic FOL systems,
to cover a wider range of input
without sacrificing the completeness guarantees and high efficiency
of today's OWL~2 DL reasoners on legal OWL~2 DL input.
As further research tasks we propose
investigating the optimization potential of FOL reasoning
for unrestricted OWL,
and determining the most relevant use cases
of non-finite model finding for OWL reasoning,
and the development of specialized reasoning methods for them.

%% file: content-acknowledgements.tex
\ACKNOWLEDGEMENTS
\label{content-acknowledgements}
Michael Schneider has been supported by the projects
\emph{SEALS} (European Commission, EU-IST-2009-238975)
and
\emph{THESEUS} (German Federal Ministry of Economics and Technology, FK~OIMQ07019).
Sebastian Rudolph is supported by the project ExpresST funded by the
German Research Foundation (DFG).

%% file: content-appx-testdata.tex
\section{Test Suite ``Complex Family Relations''}
\label{content-appx-testdata}

\input{distribution/testdata/ComplexFamilyRelations/description.txt}
The test cases are also available in electronical form
in the supplementary material.

\subsection{Test Case ``01AsymmetricTransitive''}
\label{content-appx-testdata-AsymmetricTransitive}

\input{distribution/testdata/ComplexFamilyRelations/testcases/01AsymmetricTransitive/description.txt}

\medskip
\noindent\textbf{\textsl{Modeling Pattern:}}
{\PatternAsymmetricTransitive}, Section~\ref{content-usecases-AsymmetricTransitive}.

\medskip
\noindent\textbf{\textsl{Premise:}}
\begin{quote}\TestCaseFontSize
\verbatiminput{distribution/testdata/ComplexFamilyRelations/testcases/01AsymmetricTransitive/owl2functional/premise.ofn}
\end{quote}
\noindent\textbf{\textsl{Conclusion:}}
\begin{quote}\TestCaseFontSize
\verbatiminput{distribution/testdata/ComplexFamilyRelations/testcases/01AsymmetricTransitive/owl2functional/conclusion.ofn}
\end{quote}

\subsection{Test Case ``02AsymmetricChained''}
\label{content-appx-testdata-AsymmetricChained}

\input{distribution/testdata/ComplexFamilyRelations/testcases/02AsymmetricChained/description.txt}

\medskip
\noindent\textbf{\textsl{Modeling Pattern:}}
{\PatternAsymmetricChained}, Section~\ref{content-usecases-AsymmetricChained}.

\medskip
\noindent\textbf{\textsl{Premise:}}
\begin{quote}\TestCaseFontSize
\verbatiminput{distribution/testdata/ComplexFamilyRelations/testcases/02AsymmetricChained/owl2functional/premise.ofn}
\end{quote}
\noindent\textbf{\textsl{Conclusion:}}
\begin{quote}\TestCaseFontSize
\verbatiminput{distribution/testdata/ComplexFamilyRelations/testcases/02AsymmetricChained/owl2functional/conclusion.ofn}
\end{quote}

\subsection{Test Case ``03DisjointTransitive''}
\label{content-appx-testdata-DisjointTransitive}

\input{distribution/testdata/ComplexFamilyRelations/testcases/03DisjointTransitive/description.txt}

\medskip
\noindent\textbf{\textsl{Modeling Pattern:}}
{\PatternDisjointTransitive}, Section~\ref{content-usecases-DisjointTransitive}.

\medskip
\noindent\textbf{\textsl{Premise:}}
\begin{quote}\TestCaseFontSize
\verbatiminput{distribution/testdata/ComplexFamilyRelations/testcases/03DisjointTransitive/owl2functional/premise.ofn}
\end{quote}
\noindent\textbf{\textsl{Conclusion:}}
\begin{quote}\TestCaseFontSize
\verbatiminput{distribution/testdata/ComplexFamilyRelations/testcases/03DisjointTransitive/owl2functional/conclusion.ofn}
\end{quote}

\subsection{Test Case ``04DisjointChained''}
\label{content-appx-testdata-DisjointChained}

\input{distribution/testdata/ComplexFamilyRelations/testcases/04DisjointChained/description.txt}

\medskip
\noindent\textbf{\textsl{Modeling Pattern:}}
{\PatternDisjointChained}, Section~\ref{content-usecases-DisjointChained}.

\medskip
\noindent\textbf{\textsl{Premise:}}
\begin{quote}\TestCaseFontSize
\verbatiminput{distribution/testdata/ComplexFamilyRelations/testcases/04DisjointChained/owl2functional/premise.ofn}
\end{quote}
\noindent\textbf{\textsl{Conclusion:}}
\begin{quote}\TestCaseFontSize
\verbatiminput{distribution/testdata/ComplexFamilyRelations/testcases/04DisjointChained/owl2functional/conclusion.ofn}
\end{quote}

\subsection{Test Case ``05NumberrestrictedTransitive''}
\label{content-appx-testdata-NumberrestrictedTransitive}

\input{distribution/testdata/ComplexFamilyRelations/testcases/05NumberrestrictedTransitive/description.txt}

\medskip
\noindent\textbf{\textsl{Modeling Pattern:}}
{\PatternNumberrestrictedTransitive}, Section~\ref{content-usecases-NumberrestrictedTransitive}.

\medskip
\noindent\textbf{\textsl{Premise:}}
\begin{quote}\TestCaseFontSize
\verbatiminput{distribution/testdata/ComplexFamilyRelations/testcases/05NumberrestrictedTransitive/owl2functional/premise.ofn}
\end{quote}
\noindent\textbf{\textsl{Conclusion:}}
\begin{quote}\TestCaseFontSize
\verbatiminput{distribution/testdata/ComplexFamilyRelations/testcases/05NumberrestrictedTransitive/owl2functional/conclusion.ofn}
\end{quote}

\subsection{Test Case ``06NumberrestrictedChained''}
\label{content-appx-testdata-NumberrestrictedChained}

\input{distribution/testdata/ComplexFamilyRelations/testcases/06NumberrestrictedChained/description.txt}

\medskip
\noindent\textbf{\textsl{Modeling Pattern:}}
{\PatternNumberrestrictedChained}, Section~\ref{content-usecases-NumberrestrictedChained}.

\medskip
\noindent\textbf{\textsl{Premise:}}
\begin{quote}\TestCaseFontSize
\verbatiminput{distribution/testdata/ComplexFamilyRelations/testcases/06NumberrestrictedChained/owl2functional/premise.ofn}
\end{quote}
\noindent\textbf{\textsl{Conclusion:}}
\begin{quote}\TestCaseFontSize
\verbatiminput{distribution/testdata/ComplexFamilyRelations/testcases/06NumberrestrictedChained/owl2functional/conclusion.ofn}
\end{quote}

\subsection{Test Case ``07NonregularSingleChained''}
\label{content-appx-testdata-NonregularSingleChained}

\input{distribution/testdata/ComplexFamilyRelations/testcases/07NonregularSingleChained/description.txt}

\medskip
\noindent\textbf{\textsl{Modeling Pattern:}}
{\PatternNonregularSingleChained}, Section~\ref{content-usecases-NonregularSingleChained}.

\medskip
\noindent\textbf{\textsl{Premise:}}
\begin{quote}\TestCaseFontSize
\verbatiminput{distribution/testdata/ComplexFamilyRelations/testcases/07NonregularSingleChained/owl2functional/premise.ofn}
\end{quote}
\noindent\textbf{\textsl{Conclusion:}}
\begin{quote}\TestCaseFontSize
\verbatiminput{distribution/testdata/ComplexFamilyRelations/testcases/07NonregularSingleChained/owl2functional/conclusion.ofn}
\end{quote}

\subsection{Test Case ``08NonregularMultiChained''}
\label{content-appx-testdata-NonregularMultiChained}

\input{distribution/testdata/ComplexFamilyRelations/testcases/08NonregularMultiChained/description.txt}

\medskip
\noindent\textbf{\textsl{Modeling Pattern:}}
{\PatternNonregularMultiChained}, Section~\ref{content-usecases-NonregularMultiChained}.

\medskip
\noindent\textbf{\textsl{Premise:}}
\begin{quote}\TestCaseFontSize
\verbatiminput{distribution/testdata/ComplexFamilyRelations/testcases/08NonregularMultiChained/owl2functional/premise.ofn}
\end{quote}
\noindent\textbf{\textsl{Conclusion:}}
\begin{quote}\TestCaseFontSize
\verbatiminput{distribution/testdata/ComplexFamilyRelations/testcases/08NonregularMultiChained/owl2functional/conclusion.ofn}
\end{quote}

\subsection{Test Case ``09ScopedEquivalence''}
\label{content-appx-testdata-ScopedEquivalence}

\input{distribution/testdata/ComplexFamilyRelations/testcases/09ScopedEquivalence/description.txt}

\medskip
\noindent\textbf{\textsl{Modeling Pattern:}}
{\PatternScopedEquivalence}, Section~\ref{content-usecases-ScopedEquivalence}.

\medskip
\noindent\textbf{\textsl{Premise:}}
\begin{quote}\TestCaseFontSize
\verbatiminput{distribution/testdata/ComplexFamilyRelations/testcases/09ScopedEquivalence/owl2functional/premise.ofn}
\end{quote}
\noindent\textbf{\textsl{Conclusion:}}
\begin{quote}\TestCaseFontSize
\verbatiminput{distribution/testdata/ComplexFamilyRelations/testcases/09ScopedEquivalence/owl2functional/conclusion.ofn}
\end{quote}

\subsection{Test Case ``10ReflexiveTransitive''}
\label{content-appx-testdata-ReflexiveTransitive}

\input{distribution/testdata/ComplexFamilyRelations/testcases/10ReflexiveTransitive/description.txt}

\medskip
\noindent\textbf{\textsl{Modeling Pattern:}}
{\PatternReflexiveTransitive}, Section~\ref{content-usecases-ReflexiveTransitive}.

\medskip
\noindent\textbf{\textsl{Premise:}}
\begin{quote}\TestCaseFontSize
\verbatiminput{distribution/testdata/ComplexFamilyRelations/testcases/10ReflexiveTransitive/owl2functional/premise.ofn}
\end{quote}
\noindent\textbf{\textsl{Conclusion:}}
\begin{quote}\TestCaseFontSize
\verbatiminput{distribution/testdata/ComplexFamilyRelations/testcases/10ReflexiveTransitive/owl2functional/conclusion.ofn}
\end{quote}

\subsection{Test Case ``11CyclicSingleRelation''}
\label{content-appx-testdata-CyclicSingleRelation}

\input{distribution/testdata/ComplexFamilyRelations/testcases/11CyclicSingleRelation/description.txt}

\medskip
\noindent\textbf{\textsl{Modeling Pattern:}}
{\PatternCyclicSingleRelation}, Section~\ref{content-usecases-CyclicSingleRelation}.

\medskip
\noindent\textbf{\textsl{Premise:}}
\begin{quote}\TestCaseFontSize
\verbatiminput{distribution/testdata/ComplexFamilyRelations/testcases/11CyclicSingleRelation/owl2functional/premise.ofn}
\end{quote}
\noindent\textbf{\textsl{Conclusion:}}
\begin{quote}\TestCaseFontSize
\verbatiminput{distribution/testdata/ComplexFamilyRelations/testcases/11CyclicSingleRelation/owl2functional/conclusion.ofn}
\end{quote}

\subsection{Test Case ``12CyclicMultiRelation''}
\label{content-appx-testdata-CyclicMultiRelation}

\input{distribution/testdata/ComplexFamilyRelations/testcases/12CyclicMultiRelation/description.txt}

\medskip
\noindent\textbf{\textsl{Modeling Pattern:}}
{\PatternCyclicMultiRelation}, Section~\ref{content-usecases-CyclicMultiRelation}.

\medskip
\noindent\textbf{\textsl{Premise:}}
\begin{quote}\TestCaseFontSize
\verbatiminput{distribution/testdata/ComplexFamilyRelations/testcases/12CyclicMultiRelation/owl2functional/premise.ofn}
\end{quote}
\noindent\textbf{\textsl{Conclusion:}}
\begin{quote}\TestCaseFontSize
\verbatiminput{distribution/testdata/ComplexFamilyRelations/testcases/12CyclicMultiRelation/owl2functional/conclusion.ofn}
\end{quote}

%% file: distribution/testdata/ComplexFamilyRelations/description.txt
This is the "Complex Family Relations" test suite for OWL 2 reasoning.
The test suite consists of a set of test cases that are compliant with 
the OWL 2 Structural Specification (http://www.w3.org/TR/owl2-syntax/)
and have a precise meaning under the OWL 2 Direct Semantics
(http://www.w3.org/TR/owl2-direct-semantics/), but do not conform to 
the narrower specification of OWL 2 DL (see Chapter 3 of the
OWL 2 Structural Specification). Each test case is an example 
for a particular modeling pattern, in which some of the global 
syntactic restrictions of OWL 2 DL are hurt (see Chapter 11 of the 
OWL 2 Structural Specification). Every test case consists of a 
premise and conclusion ontology, where the premise ontology is 
syntactically invalid with regard to OWL 2 DL and the conclusion 
ontology is logically entailed from the premise ontology under the 
OWL 2 Direct Semantics. The test cases represent certain family 
relationships and follow the style of the well-known "Families" 
ontology, which is used in the OWL 2 Primer for demonstrating the 
language features of OWL 2 (see Chapter 13 of the OWL 2 Primer, 
available at http://www.w3.org/TR/owl2-primer/). The syntax for 
all ontologies is the OWL 2 Functional Syntax, as defined in the 
OWL 2 Structural Specification document.

%% file: distribution/testdata/ComplexFamilyRelations/testcases/01AsymmetricTransitive/description.txt
The ancestor relationship can be represented as a 
strict partial order, that is, an asymmetric transitive 
property: If Mary has ancestor Bill who has ancestor John, 
then Mary has John as her ancestor, but John cannot 
have Mary as his ancestor. OWL 2 DL does not allow 
transitive properties to be asymmetric.

%% file: distribution/testdata/ComplexFamilyRelations/testcases/02AsymmetricChained/description.txt
The uncle relationship can be composed from the parent 
and brother properties using a property chain axiom.
Naturally, the uncle relation would further be specified 
to be asymmetric: If Mary has parent Bill who has brother 
John, then Mary has John as her uncle, but John cannot
have Mary as his uncle. OWL 2 DL does not allow composite 
properties to be asymmetric.

%% file: distribution/testdata/ComplexFamilyRelations/testcases/03DisjointTransitive/description.txt
The pair of ancestor and descendant relationships can be
represented as mutually disjoint transitive properties:
If Mary has ancestor Bill, and Bill has ancestor John, then
Mary has John as her ancestor, but not as her descendant.
OWL 2 DL does not allow to put disjointness axioms on
transitive properties.

%% file: distribution/testdata/ComplexFamilyRelations/testcases/04DisjointChained/description.txt
The uncle relationship can be composed from the parent 
and brother properties using a property chain axiom.
Naturally, all these properties would further be specified 
to be mutually disjoint: If Mary has parent Bill, and Bill 
has brother John, then Mary has John as her uncle, but 
cannot have John as her parent or brother. OWL 2 DL does 
not allow to put disjointness axioms on composite properties.

%% file: distribution/testdata/ComplexFamilyRelations/testcases/05NumberrestrictedTransitive/description.txt
The ancestor relationship between persons can be understood
to be a transitive property such that every person has
at least two ancestors: If Mary is a person, then there are
persons X and Y, such that Mary has ancestor X, person X
has ancestor Y, and Mary has ancestor Y. This can be
represented by using a minimum cardinality restriction
on a transitive property. OWL 2 DL does not allow to put
cardinality restrictions on transitive properties.

%% file: distribution/testdata/ComplexFamilyRelations/testcases/06NumberrestrictedChained/description.txt
The maternal grandfather relationship can be composed from
the mother and father properties using a property chain axiom,
where the composed property would naturally be functional: 
If Mary has mother Susan, and Susan has the two fathers Jim 
and James, then Mary has Jim and James as their maternal 
grandfathers, where Jim is identical to James.
OWL 2 DL does not allow composite properties to be functional.

%% file: distribution/testdata/ComplexFamilyRelations/testcases/07NonregularSingleChained/description.txt
The relatives relationship between two persons Susan and John
can propagate to two other persons Mary and Bill, for instance
if Mary has Susan as her mother and Bill has John as his father.
This can be expressed using property chain axioms by composing
the relative property from the mother, the relative, and the 
inverse of the father property. This representation hurts the 
regularity conditions for the property hierarchy of OWL 2 DL, 
as in chains of size 3, an inner property of the chain 
must not also occur as the composite property.

%% file: distribution/testdata/ComplexFamilyRelations/testcases/08NonregularMultiChained/description.txt
Two interlaced property chain axioms can be used to closer 
characterize the uncle relationship, by composing the uncle 
relationship from the cousin and father relationships, and 
the cousin relationship from the uncle and the inverse of the
father relationships. Hence, if Mary has cousin Bill and 
father John, and Bill has father Jim and uncle John, then 
Mary has Jim as her uncle and Bill has Mary as his cousin. 
The use of circular dependencies on the property hierarchy 
break the regularity conditions of OWL 2 DL.

%% file: distribution/testdata/ComplexFamilyRelations/testcases/09ScopedEquivalence/description.txt
Having some other person or oneself as a relative can be 
modeled as a scoped equivalence relation, for which 
application is limited to the class of persons, i.e., 
as a symmetric transitive property with a self-restriction 
over the Person class. Hence, if Mary has relatives John 
and Jim and Bill is another person, then John has relatives 
Mary and Jim, and Mary and Bill each have themselves as 
relatives. However, if C3PO is not a person, then C3PO 
cannot have itself as a relative. OWL 2 DL does not allow 
for self-restrictions on transitive properties.

%% file: distribution/testdata/ComplexFamilyRelations/testcases/10ReflexiveTransitive/description.txt
The reflexive-transitive closure of the parent relationship 
between persons is the extended ancestor relationship
that is also reflexive, while being scoped to persons. 
It can be coarsely approximated by a transitive super property 
of the parent property that is used in a self-restriction 
over the class of persons. Hence, if Mary has parent Bill 
who has parent John, then Mary is in the extended ancestor
relationship with Bill, John and herself. However, if C3PO 
is not a person, then C3PO cannot be in an ancestor 
relationship with itself. OWL 2 DL does not allow for 
self-restrictions on transitive properties.

%% file: distribution/testdata/ComplexFamilyRelations/testcases/11CyclicSingleRelation/description.txt
Binary relationships, such as loving someone, may be used 
to build circular relationships and the relationship cycles 
may be of arbitrary size: Jack may only love himself, 
John and Joan may love each other, and Mary may love Bill, 
who loves Susan, who loves Jim, who again loves Mary. 
Persons being in such a loves-relationship cycle can be 
seen as instances of the class of loved lovers. In our 
examples, all listed persons would be instances of this 
class. Class instanceship can be expressed in terms of 
a self-restricted transitive super property of the loves 
property, but this is not allowed in OWL 2 DL.

%% file: distribution/testdata/ComplexFamilyRelations/testcases/12CyclicMultiRelation/description.txt
A legitimate child is a person with a father and a mother 
who are married. Hence, this relation is composed from a set 
of three different basic relations in a cyclic way. Occurrence 
of such relationships in a knowledge base can be indicated by 
instanceship in a class of legitimate children, modeled using 
a property chain axiom and a self-restriction. OWL 2 DL does 
not allow self-restrictions on composite properties.

%% file: content-appx-results.tex
\section{Detailed Results}
\label{content-appx-results}

\input{distribution/results/description.txt}
The results are also available in electronical form
in the supplementary material.

\subsection{OWL 2 DL Reasoners: Entailment Checking Results}
\label{content-appx-results-owldlentail}

\input{distribution/results/owldlentail/description.txt}

\subsubsection{Fact++}
\label{content-appx-results-owldlentail-factpp}
\RawResult{distribution/results/owldlentail/rawresults-complexrel-owldl-entail-factpp.txt}

\subsubsection{HermiT}
\label{content-appx-results-owldlentail-hermit}
\RawResult{distribution/results/owldlentail/rawresults-complexrel-owldl-entail-hermit.txt}

\subsubsection{Pellet}
\label{content-appx-results-owldlentail-pellet}
\RawResult{distribution/results/owldlentail/rawresults-complexrel-owldl-entail-pellet.txt}

\subsection{FOL Reasoners: Entailment Checking Results}
\label{content-appx-results-folatpentail}

\input{distribution/results/folatpentail/description.txt}

\subsubsection{E}
\label{content-appx-results-folatpentail-e}
\RawResult{distribution/results/folatpentail/rawresults-complexrel-folatp-entail-e.txt}

\subsubsection{iProver}
\label{content-appx-results-folatpentail-iprover}
\RawResult{distribution/results/folatpentail/rawresults-complexrel-folatp-entail-iprover.txt}

\subsubsection{SPASS}
\label{content-appx-results-folatpentail-spass}
\RawResult{distribution/results/folatpentail/rawresults-complexrel-folatp-entail-spass.txt}

\subsubsection{Vampire}
\label{content-appx-results-folatpentail-vampire}
\RawResult{distribution/results/folatpentail/rawresults-complexrel-folatp-entail-vampire.txt}

\subsubsection{DarwinFM}
\label{content-appx-results-folatpentail-darwinfm}
\RawResult{distribution/results/folatpentail/rawresults-complexrel-folatp-entail-darwinfm.txt}

\subsubsection{Paradox}
\label{content-appx-results-folatpentail-paradox}
\RawResult{distribution/results/folatpentail/rawresults-complexrel-folatp-entail-paradox.txt}

\subsection{FOL Reasoners: Satisfiability Checking Results}
\label{content-appx-results-folatpsat}

\input{distribution/results/folatpsat/description.txt}

\subsubsection{E}
\label{content-appx-results-folatpsat-e}
\RawResult{distribution/results/folatpsat/rawresults-complexrel-folatp-sat-e.txt}

\subsubsection{iProver}
\label{content-appx-results-folatpsat-iprover}
\RawResult{distribution/results/folatpsat/rawresults-complexrel-folatp-sat-iprover.txt}

\subsubsection{SPASS}
\label{content-appx-results-folatpsat-spass}
\RawResult{distribution/results/folatpsat/rawresults-complexrel-folatp-sat-spass.txt}

\subsubsection{Vampire}
\label{content-appx-results-folatpsat-vampire}
\RawResult{distribution/results/folatpsat/rawresults-complexrel-folatp-sat-vampire.txt}

\subsubsection{DarwinFM}
\label{content-appx-results-folatpsat-darwinfm}
\RawResult{distribution/results/folatpsat/rawresults-complexrel-folatp-sat-darwinfm.txt}

\subsubsection{Paradox}
\label{content-appx-results-folatpsat-paradox}
\RawResult{distribution/results/folatpsat/rawresults-complexrel-folatp-sat-paradox.txt}

%% file: distribution/results/description.txt
These are the detailed results of all the evaluations
that have been conducted for OWL 2 DL reasoners 
(entailment tests) and FOL reasoners (entailment and 
satisfiability tests) using the Complex Family Relations
test suite.

%% file: distribution/results/owldlentail/description.txt
These are the detailed entailment checking results for the 
OWL 2 DL reasoners using the twelve entailments of the 
"Complex Family Relations" test suite. For each reasoner, 
there is a list of reasoning results, where each entry is 
a section telling 
    the name of the test case,
    the type of the test case (always "entailment checking"),
    all output written by the reasoner (including error messages),
    the reasoning outcome, and
    the CPU time in milliseconds (ms).
Possible reasoning outcomes are:
    "entailment" (correct result),
    "non-entailment" (wrong result),
    "timeout" (timeout), and
    "error" (error).

%% file: distribution/results/folatpentail/description.txt
These are the detailed entailment checking results for the 
FOL reasoning systems using the twelve entailments of the 
"Complex Family Relations" test suite. Both theorem provers 
and model finders were tested. For each reasoner, there is 
a list of reasoning results, where each entry is a line telling 
    the number of the test case, 
    the name of the reasoner,
    the version of the reasoner, 
    the reasoning outcome, and
    the CPU time in seconds.
Possible reasoning outcomes are:
    "Theorem" (correct result),
    "CounterSatisfiable" (wrong result),
    "Timeout" (timeout), and
    "Error" (error; other error messages are possible).

%% file: distribution/results/folatpsat/description.txt
These are the detailed satisfiability checking results for the 
FOL reasoning systems using the twelve premise ontologies of 
the "Complex Family Relations" test suite. Both theorem provers 
and model finders were tested. For each reasoner, there is 
a list of reasoning results, where each entry is a line telling 
    the number of the test case, 
    the name of the reasoner,
    the version of the reasoner, 
    the reasoning outcome, and
    the CPU time in seconds.
Possible reasoning outcomes are:
    "Satisfiable" (correct result),
    "Unsatisfiable" (wrong result),
    "Timeout" (timeout), and
    "Error" (error; other error messages are possible).

%% file: content-appx-owl2tptp.tex
\section{Tool for Translating OWL 2 Ontologies into TPTP}
\label{content-appx-owl2tptp}

In this section,
we give an overview to our tool
for translating OWL 2 ontologies into FOL (TPTP),
which implements the approach given in 
Section~\ref{content-folapproach-approach},
and which was used in the evaluation in 
Section~\ref{content-folapproach-evaluation}.
The executable tool and its source code
are available in electronical form
in the supplementary material.

\medskip
\hrule
\medskip

{\small
\begin{verbatim}
owldirect2tptp: A tool to translate OWL 2 ontologies or queries
into TPTP-style FOL formulae according to the OWL 2 Direct Semantics.
2012 by Michael Schneider <schneid@fzi.de>

USAGE: 

  java -jar owldirect2tptp.jar <Command> [<Options>...]
    * plain <OntologyLoc> <OutputLoc> [<Semantics>]
    * axiom <OntologyLoc> <FormulaName> <OutputLoc> [<Semantics>]
    * conjecture <OntologyLoc> <FormulaName> <OutputLoc> [<Semantics>]
    * question <OntologyLoc> <FormulaName> <OutputLoc> [<Semantics>]
    * check-unsupported <OntologyLoc> [<Semantics>]

All ontologies are read from the specified input locations. The input
format can be any valid OWL 2 encoding (e.g. OWL 2 Functional Syntax,
RDF/XML, or Manchester Syntax). The resulting TPTP formulae will have
the formula names that are specified as arguments. The result is
written into a single document, for which the output location
is defined as an argument. The semantics underlying the translation
into FOL can be optionally defined as either the
standard first-order-based OWL 2 Direct Semantics ('standard')
or a HiLog-style variant of it ('hilog'). The default semantics
is 'standard'.

If an unsupported language construct is being used in an axiom
of an ontology, the axiom is ignored by the translation. Use the
command 'check-unsupported' to find out whether an ontology
contains any unsupported features, in which case the return value
will not be 0 and a list of messages is printed indicating the issues.

For more details on the translation as well as on current limiations,
see the description of package
org.swertia.plugins.translation.standard.owldirect.essentials.translation.tptp .

Note: To avoid out-of-memory errors when translating large ontologies,
set the heap space of the Java runtime environment to a sufficient size
via option '-Xms' (e.g. 'java -Xms1500M' for 1.5GB RAM).
\end{verbatim}
}